\pdfoutput=1

\documentclass[11pt]{article}

\usepackage[final]{acl}

\usepackage{times}
\usepackage{latexsym}

\usepackage[T1]{fontenc}

\usepackage[utf8]{inputenc}

\usepackage{microtype}

\usepackage{inconsolata}

\usepackage{graphicx}
\usepackage{amsmath}
\usepackage{tabularray}
\usepackage{svg}
\usepackage[export]{adjustbox}
\graphicspath{{images/}}

\title{How and where does CLIP process negation?}

\author{Vincent Quantmeyer\thanks{Work carried out as M.Sc.\@ student at Utrecht University} \\
  Department of Information\\ and Computing Sciences \\
  Utrecht University\\
  \texttt{v.quantmeyer@gmail.com}\\\And
  Pablo Mosteiro \\
  Department of Methodology\\ and Statistics \\
  Utrecht University \\
  \texttt{p.mosteiro@uu.nl} \\\And
  Albert Gatt \\
  Department of Information\\ and Computing Sciences \\
  Utrecht University \\
  \texttt{a.gatt@uu.nl} \\
  }

\begin{document}
\maketitle
\begin{abstract}
Various benchmarks have been proposed to test linguistic understanding in pre-trained vision \& language (VL) models.
Here we build on the existence task from the VALSE benchmark \citep{parcalabescu2022} which we use to test models' understanding of negation, a particularly interesting issue for multimodal models. 
However, while such VL benchmarks are useful for measuring model performance, they do not reveal anything about the internal processes through which these models arrive at their outputs in such visio-linguistic tasks.
We take inspiration from the growing literature on model interpretability to explain the behaviour of VL models on the understanding of negation.
Specifically, we approach these questions through an in-depth analysis of the text encoder in CLIP 
\citep{radford2021}, a highly influential VL model. We localise parts of the encoder that process negation and analyse the role of attention heads in this task.
Our contributions are threefold. We demonstrate how methods from the language model interpretability literature (such as causal tracing) can be translated to multimodal models and tasks; we provide concrete insights into how CLIP processes negation on the VALSE existence task; and we highlight inherent limitations in the VALSE dataset as a benchmark for linguistic understanding.
\end{abstract}

\section{Introduction}

Research in vision \& language (VL) modelling has produced various pre-trained models that are capable of jointly processing image and text information by learning multimodal representations \citep[e.g.,][]{li2019,lu2019,radford2021,jia2021,li2021}.
This makes them applicable to a host of downstream tasks, such as visual question answering, image caption generation or zero-shot image classification.

\begin{figure*}
    \centering
    \begin{tblr}{colspec={l l X X},
        row{1} = {font=\bfseries},
        hline{1,Z}={1.5pt},
        hline{2-Y}={0.8pt}}

        Type & Image & Caption & Foil\\

        {Negation\\in foil} & \includegraphics[scale=0.2, valign=c]{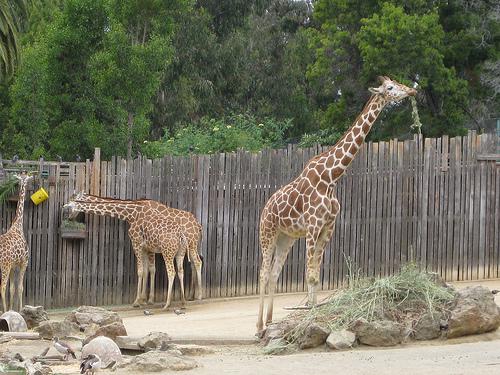} & \textrm{There are giraffes} & \textrm{There are no giraffes}\\

        {Negation\\in caption} & \includegraphics[scale=0.2, valign=c]{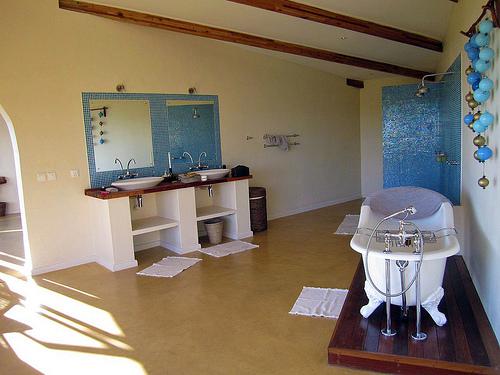} & \textrm{There are no people} & \textrm{There are people}\\

    \end{tblr}
    \caption[Examples from VALSE existence]{Examples from VALSE existence \citep{parcalabescu2022}. 
    Caption and foil only differ in the presence or absence of the negator ``no''.
    The negator is either in the caption or the foil.}
    \label{fig:valse_existence}
\end{figure*}

Various benchmarks have been proposed to test these models' understanding of different linguistic features, such as word order \citep{akula2020}, verb meaning \citep{hendricks2021}, and compositionality \citep{thrush2022}.
The VALSE benchmark \citep{parcalabescu2022} was introduced to test these models' ability to ground features such as existence, plurality, or spatial relations in images. 
An example of the existence piece is shown in Figure \ref{fig:valse_existence}. Given an image, a model must choose between a correct caption and an incorrect foil, one of which contains a negation operator.

As such, this piece can be used to test a model's understanding of negation, a particularly interesting issue for multimodal models, which typically include a visual backbone pretrained on computer vision tasks such as object labelling. The models themselves are further pretrained on image-text pairs where there is likely to be a {\em positive} bias, since captions describing images will typically refer to what is depicted there. This raises the question whether VL models are capable of processing operators such as ``no'' in instances such as those in Figure \ref{fig:valse_existence}. Indeed, negation remains a weakness of even the most state-of-the-art large language models \citep{truong2023}

In line with these intuitions, initial VALSE results reveal that models only achieve moderate performance in this (and other) linguistic categories. 
However, while VL benchmarks such as VALSE are useful for measuring current and future model performance, they do not reveal anything about the internal processes through which these models arrive at their outputs in such visio-linguistic tasks.

We aim to make use of the growing literature on model interpretability \citep{rauker2022} in order to explain the behaviour (and shortcomings) of VL models on 
the understanding of negation. To do this, we use the existence sub-task in VALSE, with some extensions, 
exploiting localisation techniques to quantify the roles that different model components play in this task. 
This yields the following research question:
{\em Which components of VL models are responsible for the model's understanding of negation?} 
We address two issues that arise from this general question, namely (1) the extent to which processing of negation is localised vs. distributed; and (2) whether model performance on VALSE-like tasks involving negation can in part be explained by high-level dataset features.

Specifically, we approach these questions through an in-depth analysis of CLIP \citep{radford2021}, a highly influential VL model. CLIP has a relatively simple design based exclusively on Transformers, which allows us to leverage interpretability techniques that target this architecture. Additionally, prior work by \citet{parcalabescu2023} shows that CLIP makes balanced use of text and image input and avoids so-called unimodal collapse \citep{madhyastha_defoiling_2018,hessel2020,Frank2021}, an important consideration for a study of multimodal model interpretability. 
Finally, CLIP remains central to developments in both vision \citep[e.g. the CLIPSeg segmentation model;][]{luddecke2022} and VL tasks \citep[e.g. CLIP is a component of several text-to-image and image-to-text models, including][among others]{mokady2021,li2023,ramesh2022,rombach2022}.

In our analysis of negation, we focus on the CLIP text encoder. However, it is important to note that CLIP is pretrained with a multimodal contrastive objective, which has been shown to yield different representations compared to text-only encoders with comparable architecture but different training objectives~\citep{wolfe2022}. Thus, we take the insights into the text encoder's ability to process negation as reflecting on the success or otherwise of the contrastive, multimodal pretraining in such models.

The contributions of this work are threefold: firstly, we demonstrate how methods from the language model interpretability literature \citep[e.g., causal tracing;][]{meng2023} can be translated to multimodal models and tasks; secondly, we provide concrete insights into how CLIP processes negation on the VALSE existence task; thirdly, we highlight inherent limitations in the VALSE dataset as a benchmark for linguistic understanding.

\section{Related work}
\paragraph{Vision-and-language models}
VL pretraining gained impetus from the development of multimodal, pretrained encoders inspired by BERT \cite{devlin2019}. \citet{bugliarello2021} provide a unified analysis of the varying VL BERT architectures.
%

With the introduction of CLIP \citep{radford2021}, contrastive learning objectives have become prominent in VL models, with or without additional objectives that address multimodal fusion \cite{jia2021,li2021,singh2022,zeng2022}. Models such as BLIP \citep{li2022} and FLAVA \citep{singh_flava_2022} combine contrastive objectives with unimodal pretraining of vision and language encoders. Architectures such as Flamingo \citep{alayrac2022} and BLIP-2 \citep{li2023} reduce training cost by training relatively small networks to map between representations from pretrained image and language models.

In CLIP,
an image encoder and a text encoder process their respective inputs completely separately from each other, i.e., without any multimodal cross-attention
and 
project them into the same latent space. The goal of contrastive learning is to maximise similarity between matching image-text pairs, minimising the similarity between non-matching pairs.
During inference, CLIP 
computes the similarity of an image and a text in the form of a 
scaled dot product between their embeddings.
Contrastive objectives have been shown to yield better embedding representations \citep{wolfe2022} leading to improved performance on semantic evaluation tasks \citep{mu2018}.

\paragraph{Vision-and-language benchmarks}

VL benchmarks focusing on specific linguistic phenomena play an important role in highlighting strengths and weaknesses in models' grounding capabilities. For example, a recent study combining several benchmarks \citep{bugliarello2023} showed that models still find certain linguistic phenomena challenging, and that grounding capabilities may be less related to model size, and more to other variables, including the fine-grained object recognition capabilities of the visual backbone \citep[e.g.][]{zheng2022}.

One class of benchmarks focuses on the robustness of models to syntactic permutations and/or their ability to reason compositionally when predicting whether visual inputs correspond to linguistic descriptions \citep[e.g.,][]{akula2020, hendricks2021,thrush2022,ma2023,yuksekgonul2023,chen_bla_2023}. Some of these benchmarks focus on specific linguistic phenomena, such as spatial relations \citep{liu2023,kamath_whats_2023} or temporal relations \citep[e.g.][]{kesen2024vilma}.


VALSE \citep{parcalabescu2022}, on which we build the present study, 
prompts a model with an image along with both its correct caption and a foiled caption and tests a model's ability to distinguish the caption from foil. This extends the original foiling task introduced by \citet{shekhar2017}.
VALSE is divided into six sub-tasks or `pieces', corresponding to six different linguistic phenomena. In this paper, we focus exclusively on the existence piece; see Figure \ref{fig:valse_existence}.


\paragraph{Model interpretability} 
\citet{rauker2022} define inner interpretability methods as those that help understand a model's internal structures and activations. 
One recurring strategy in such techniques is to analyse the effect of perturbations or ablations on the model's behaviour and output, whether this is applied to individual neurons \cite[e.g.][]{zhou2018,ghorbani2020} or to weights, with the goal of identifing modular subnetworks \citep{csordas2021}.


The choice of a suitable level of granularity at which to apply ablation is largely dictated by the model's size and complexity. 
Interpretability methods for transformers often operate at the level of attention heads, MHA modules, MLPs, or 
full Transformer layers.\footnote{\citet{goh2021} also produced neuron-level interpretations of CLIP's image encoder, albeit the ResNet and not the ViT variant.}
\citet{meng2023} introduced the causal tracing methodology to localise factual associations in a model. 
In \citet{meng2023}, this localisation step serves as the basis for subsequent model editing in the ROME method. 
However, follow-up work has suggested that 
the ability to edit knowledge in a particular layer does not imply that this knowledge is localised in this layer~\citep{hase2023} and can also introduce unwanted side effects \citep{hoelscher-obermaier2023}.
Given these uncertainties surrounding model editing techniques, the present study focuses on localisation only. 

A final line of relevant interpretability literature 
focuses on 
attention patterns in large Transformer models, which reveal the role of specific attention heads in processing linguistic phenomena such as syntactic roles \citep{clark2019,kovaleva2019,vig2019}. 
All of these studies converge on the finding that pre-trained Transformer language models allocate significant attention to tokens that do not carry inherent semantic meaning, such as the separator token in BERT or the start-of-sequence token in GPT-2. 

\section{Methods}
\subsection{Definitions}

A forward pass in CLIP of a single VALSE existence instance (Fig. \ref{fig:valse_existence}) consists of a text caption, a text foil, and an image. 
This produces one similarity score for caption and image and one for caption and foil, 
denoted $S_{c,i}$ and $S_{f,i}$, respectively. 

CLIP is said to correctly classify a caption-foil-image triple if $S_{c,i} > S_{f,i}$. 
We can quantify CLIP's classification performance using the difference between the two similarities. 
We denote this classification score $d = S_{c,i} - S_{f,i}$ and the absolute size of $d$ can be seen as an indicator of CLIP's confidence in the classification. 

\subsection{Data}

The VALSE existence benchmark consists of $505$ image-caption-foil triples.
The dataset is divided into instances where the negation is in the foil ($249$) and instances where the negation is in the caption ($256$). The presence or absence of a negation operator means that sometimes captions or foils can differ in token length. For our purposes, it is important that strings are of equal length; hence we insert the word \emph{some} before the noun in non-negated sentences.
See Appendix \ref{app:rephrasing} for full details.

CLIP only achieves a moderate accuracy of $0.686$ on VALSE existence. 
To 
identify patterns of processing in the model that give rise to correct classification of negation 
it is necessary to analyse correctly and incorrectly classified instances separately.
To do this consistently
across different analyses, the dataset was divided into three segments (correct, ambiguous, incorrect) based on the classification score $d$. Table \ref{tab:segmentation} shows the distribution of instances per segment.

\begin{table}[!t]
    \centering
\begin{tblr}
    {width=\textwidth,
        colspec={l r r r},
        row{1} = {font=\bfseries},
        column{1} = {font=\bfseries},
        cell{2-Z}{2-Z} = {mode=math},
        hline{1,Z} = {1.5pt},
        hline{3} = {0.8pt}}

    &  Correct & Ambiguous & Incorrect \\
    & d > 1 & 1 \geq d > -1 & d \leq -1 \\ 
    Caption & 72 & 150 & 28 \\
    Foil & 81 & 145 & 14 \\
\end{tblr}
\caption{Number of instances per segment in the VALSE existence dataset.}
    \label{tab:segmentation}
\end{table}

\subsection{Causal tracing}

\begin{figure}[!t]
    \centering
    \includegraphics[scale=0.4]{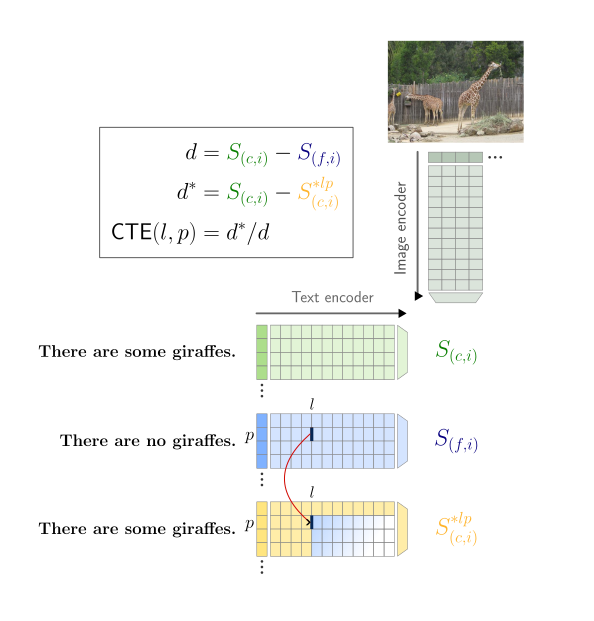}
    \caption{
        Illustration of the causal tracing methodology. 
        The activation at a single position and layer from the negated forward pass are inserted into the corresponding layer and position of the non-negated forward pass.
        This shows what proportion of the original effect can be restored by this layer-position pair.
        Image and text are taken from VALSE existence \citep{parcalabescu2022}.
        }
    \label{fig:causal_tracing_fig}
\end{figure}

Here we outline our adaption of the causal tracing method from \citet{meng2023} for the part of the dataset where the negation is in the foil. 
Figure~\ref{fig:causal_tracing_fig} provides a visual summary of the method. 

A standard forward pass is carried out with caption, foil, and image, yielding the regular classification score $d = S_{c,i} - S_{f,i}$. 
Importantly, the activations from the forward pass at each layer and each position in the text encoder are recorded. 
In the subsequent modified forward pass only the (non-negated) caption 
is used in the forward pass alongside the image. 
During this forward pass, the text encoder's activation at a given layer and position is replaced by the activation from the foil's original forward pass at the corresponding layer and position.
This is done individually for each combination of layer and position. 

Intuitively, this achieves the following. The 
model processes the non-negated caption, but at a given layer and position it is made to behave as if it was processing the negated foil. 
If, and only if,
a certain layer and position is specialised in processing negation, then substituting the activation from the negated forward pass into the non-negated one should 
affect the output in a visible way.

This intuition is quantified in the following way. 
For a given layer $l$ and a position $p$ the modified forward pass produces a similarity score $S_{c,i}^{*lp}$. 
This allows us to calculate a modified classification score $$d^* = S_{c,i} - S_{c,i}^{*lp}$$
With this modified classification score 
we calculate the causal tracing effect of layer $l$ at position $p$ 
$$\text{CTE}(l,p) = d^* / d$$ 
This effect represents the proportion of the original classification score $d$ that can be ``restored'' by layer $l$ at position $p$. 


To apply this method to cases where the negation is in the caption, one has to swap caption and foil such that, once again, the activations from the negated sentence (now the caption) are substituted into the forward pass of the non-negated sentence (now the foil).
This means that we obtain a modified classification score, which is used to calculate the causal tracing effect in the same way. $$d^{*} = S_{f,i}^{*lp} - S_{f,i}$$ 

This method yields a causal tracing effect for each layer and position for each VALSE existence triple. 
All captions in the dataset share the same beginning (SOT, There, is/are, a/some, subject) and ending set of tokens (., EOT). 
However, they differ in the number of tokens in between these two sets. 
Therefore, the CTE from all positions in between the beginning and end sets of tokens are averaged into one placeholder position called ``further subject tokens''. 
If there are no positions between the beginning and end sets, then a CTE of 0 is recorded at this position. 
Consequently, we can average CTEs across the dataset (or a segment thereof).
To represent each instance according to its sequence length, the averaged effect at the ``further subject tokens'' position is weighted by the number of tokens that make up this position in each instance. 

Lastly, we want to be able to describe the degree of localisation in particular layers.
Localisation is strongest when one position in a layer, to the exclusion of all other positions,  restores the full effect.
Conversely, localisation is absent when each position restores the same proportion of the effect.
Hence, we can quantify the degree of localisation in a layer $l$ as the standard deviation of the causal tracing effects at each position in this layer, starting at the negator position.

\subsection{Negator-selective attention in text encoder}

The purpose of this analysis is to identify attention heads in CLIP's text encoder that selectively pay attention to negators. 
%
Since a regular forward pass consists of both caption and foil, this yields two attention maps per head in the text encoder. 
Each attention map is an array of size $P \times P$ where $P$ is the number of positions in the input sequence, where the attention mask forces all elements to the right of the diagonal of this array to be $0$. 

The attention map is filtered to the column representing the position of the negator in the negated input sentence (or the quantifier in the corresponding non-negated sentence).
To identify negator-selective attention, we subtract the values from the non-negated sentence from those from the negated sentence.
Finally, the maximum of the resulting difference values is taken over all source positions and this represents the amount of negator-selective attention of a particular attention head on this particular dataset instance. 
This procedure can then be repeated over the whole dataset yielding an average negator-selective attention value $a_{lh}^N$ for each attention head $h$ in each layer $l$.

Instead of taking the maximum value over source positions, negator-selective attention can also be calculated for each source position.
In heads with high negator-selective attention, this creates a more fine-grained picture of the negator-selective attention patterns involved. 


To test the validity of the results from this analysis, it is further adapted to a subset of the CANNOT dataset \citep{anschutz2023},
from which we use $554$ negated sentences and create a positive counterpart for each. See Appendix \ref{app:cannot} for details.

\section{Results}

\subsection{Causal tracing in text encoder} \label{sec:result_causal_tracing}

\begin{figure*}
    \centering
    \includegraphics[width=0.95\textwidth]{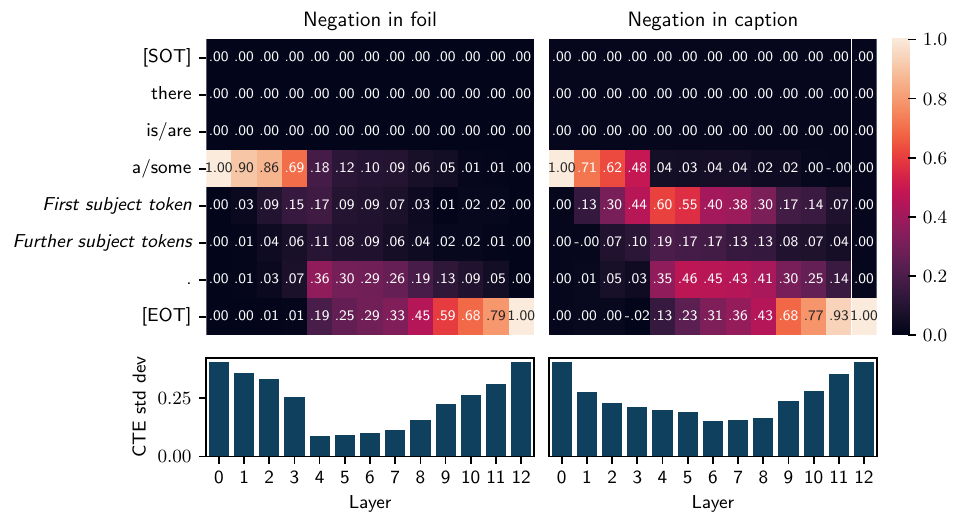}
    \caption[Causal tracing results]{
        Causal tracing effect (CTE) of the correct segment, split by whether negation is in foil or caption.
        The heatmaps show the CTE of each layer-position pair in the text encoder. 
        The bar charts show the standard deviation of all CTE in the corresponding layer as an overall measure of localisation.
        Layer 0 denotes the embedding layer.}
    \label{fig:causal_tracing}
\end{figure*}

The left heatmap in Figure \ref{fig:causal_tracing} shows the causal tracing effect per layer and position for the correct segment of the data 
with negation 
in the \emph{foil}. 


We are interested in the effect of components that lie in between the negator position in layer~0 (embeddings) and the last position in the final layer (encoder output), as these possibly mediate CLIP's correct processing of negation in the text input.\footnote{Since the encoder uses masked attention, positions prior to the negator position cannot be affected by the intervention and therefore do not show any effect.}
Figure \ref{fig:causal_tracing} shows that this effect is limited to only a subset of positions and layers and seems to suggest a path through the model.
In particular, in layers~0-3 the effect is practically limited to the negator position, suggesting that in these early layers the negation information is processed mainly at its original position. 
The effect at the negator position then drops sharply at layer~4 and further decreases until the final layer.
This indicates that the negator position only plays a pivotal role in the early layers and that the processing is in fact shifted to the second-to-last and last positions at layer~4. 
We will return to this 
in the analysis of attention patterns in Section \ref{sec:result_attention}.
In the central layers~4-7 these two positions seem to play an equally important role, judging by their respective CTE, and from layer~8 onwards, the effect is concentrated in the last layer. 

The bar charts in Figure~\ref{fig:causal_tracing} show the degree of localisation in each layer asmeasured by the standard deviation of the CTE.
In line with the interpretation above, localisation is high in the early layers~0-3, then drops sharply in layer~4, remains low in the middle layers, and goes up again in the late layers~9-12. 

The right part of Figure~\ref{fig:causal_tracing} shows the results from the same experiment on the correct segment of the data where the negation is in the \emph{caption}.
The general pattern of these results is comparable to the one
described above.
However, the first subject position already has a visible effect in the early layers, leading to reduced localisation. 
The effect of the first subject position becomes most pronounced in the middle layers which constitutes the most substantial difference between the two sets of results and in fact leads to greater localisation in the middle layers. 
In the late layers 9-12, the effect is once again concentrated in the last position. 

\subsection{Negator-selective attention in text encoder} \label{sec:result_attention}

\begin{figure*}
    \centering
    \includegraphics[width=0.8\textwidth]{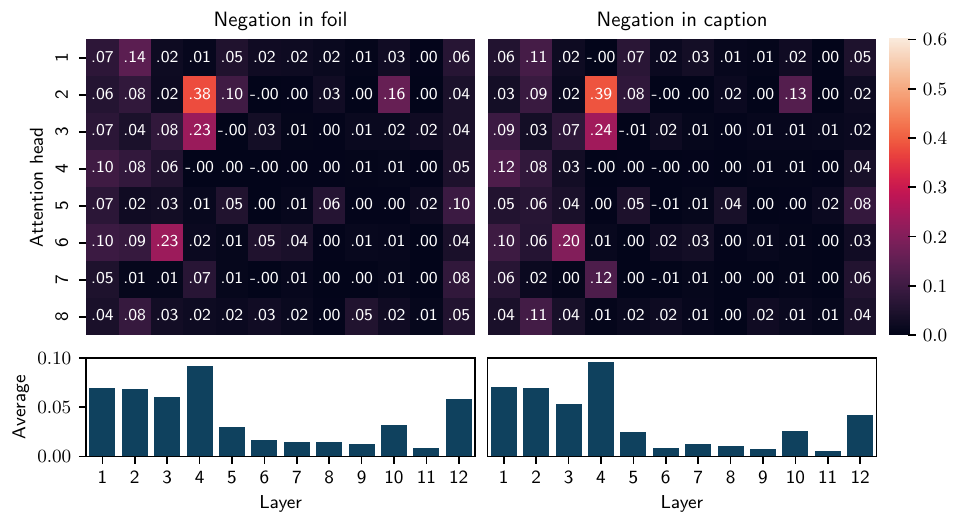}
    \caption[Negator-selective attention results]{
        Negator-selective attention across all dataset segments, split by whether negation is in foil or caption.
        The heatmaps indicate the degree of negator-selective attention for each attention head and layer.
        The bar charts show the average of each layer as an overall measure of negator-selective attention.
        }
    \label{fig:attention_2in1}
\end{figure*}

Figure \ref{fig:attention_2in1} shows the negator-selective attention of each attention head of each layer in CLIP's text encoder, divided by whether the negation is in foil or caption. 
As expected, the patterns in both parts of the dataset are practically identical, since this analysis is not affected by any visual input.
As a general observation, only a small subset of heads display any negator-selective attention ($8\% \text{ of heads with } a_{lh}^N > 0.1$) and the majority of them are found in the early layers.
The most negator-selective attention head is found in layer 4. 

Note that these results are reported across all dataset segments (incorrect, ambiguous, correct), since the patterns do not meaningfully differ between them. 
This suggests that negator-selective attention cannot explain the difference in CLIP's classification performance on different instances of VALSE existence, since the same patterns are found in correctly and incorrectly classified cases. 
In fact, none of the attention heads that show negator-selective attention of at least $0.1$ show a correlation between negator-selective attention and classification score (all $|r| < 0.2$). 

Layer 4, where the most negator-selective attention is found, is the same layer, where the causal tracing results from Section \ref{sec:result_causal_tracing} suggested that negation information is moved from its original position to later positions, in particular the second-to-last one. 
We analyse the source of this negator-specific attention, i.e., which specific positions attend particularly to the negator position in the identified heads of interest. 
Figure \ref{fig:attention_layer_4} (Appendix \ref{app:appendix_attention}) confirms that the source of negator-selective attention in Head 2 is the second-to-last position.
Furthermore, when the negation is in the caption, we find that additional negator-selective attention comes from the first subject position, which aligns with the greater role this position plays in this part of the dataset, as already suggested by the causal tracing results in Section \ref{sec:result_causal_tracing}.
Thus, the causal tracing and negator-selective attention results form a coherent narrative.

We validate these observations using the 
CANNOT dataset. Here, we observe similar trends, with most negator-selective attention found in the early layers~1-4. See Appendix \ref{app:cannot-results} for details.

\subsection{Dataset features} \label{sec:results_dataset}
We investigate whether the similarity between a caption and a foil for a given VALSE instance is correlated with the instance's classification score. Full details are in Appendix \ref{app:data_feat}, especially Figure~\ref{fig:score_vs_similarity}. We make two primary observations. First the classification score is weakly correlated with the similarity between caption and foil, especially for those instances when the negation is in the foil. Second, longer sequences exhibit greater foil-caption similarity, leading to lower scores.

\begin{figure}[!t]
    \centering
    \includegraphics[scale=0.5]{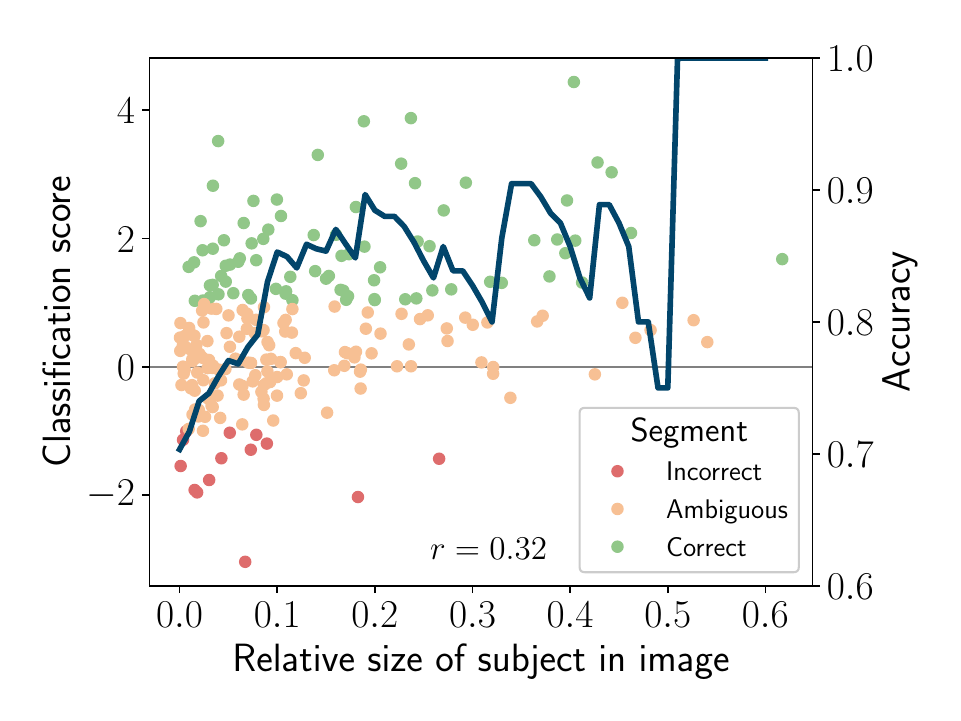}
    \caption[Subject size vs. CLIP's classification score]{
        Relative size of image subject vs. CLIP's classification score.
        All instances where the subject from the caption is shown in the image.
        Colour indicates dataset segment.
        The blue line shows classification accuracy when imposing a minimum subject size threshold.
        }
    \label{fig:score_vs_size}
\end{figure}

To investigate the effect of the size of the caption's subject (e.g. `giraffe' in Figure~\ref{fig:valse_existence}), we find its bounding box using CLIPSeg \citep{luddecke2022}, and compare its relative size to the instance's classification score (Figure~\ref{fig:score_vs_size}).
The correlation of $r = 0.32$ shows that images with more prominent subjects tend to be classified more accurately. 
In fact, when imposing a subject size threshold of $0.1$ (which removes $43\%$ of instances), CLIP achieves an accuracy of $0.85$. 
The accuracy as a function of the subject size threshold is shown by the line in Figure \ref{fig:score_vs_size}. 
Note, however, that the validity of these results decreases with higher thresholds, as the remaining sample size gets very small. 
Nonetheless, these results suggest that CLIP  exhibits better existence classification results on instances with more salient subjects.

\section{Discussion}

The causal tracing results from Section \ref{sec:result_causal_tracing} suggest relatively strong localisation in the early (1-3) and late (8-12) layers, meaning that negation is largely represented at singular positions in these layers.

In layer 4, the CTE at the negator position drops sharply, 
and this conincides 
with the finding of negator-selective attention heads in layer 4 which appear to shift negation information to later positions. 
The locations of these attention heads also overlap with those found on the CANNOT dataset, which provides initial evidence that the CLIP text encoder uses certain attention heads for specific syntactic functions. 


In the middle layers localisation is generally lower
with no single position restoring more than $60\%$ of the original effect. 
This implies that representation of negation is distributed across positions
and that the model relies on \emph{combining} the representations at each position in order to make correct judgements about negations.



Furthermore, the first subject token position appears to play a unique role in cases with
negation 
in the caption, which could be due to the asymmetry in the two tasks. 
When the negation is in the foil, the label's subject is shown in the image and, intuitively, once it is detected, a decision can be made and no further processing is necessary. 
Conversely, when the negation is in the caption, the entire image needs to be scanned to ensure that the label's caption is in fact absent from all parts of the image. 
This difference 
could be part of the reason why the first subject token position appears to play a role up until deeper layers of the network, when the negation is in the caption. 
The effects of the subject position in deeper layers could imply that the subject information is in fact more deeply processed and thus more strongly represented in the final text encoder's output which, in turn, could be conducive to the model's task of ``searching'' for the subject in the image's representation. 
However, 
these explanations are speculative and must not be accepted without further experiments. 




Section \ref{sec:results_dataset} highlights
that
the label's length and the subject's size in the image show non-negligible correlations with respect to the classification score. 
This suggests that CLIP is better at the VALSE Existence task when labels are shorter and therefore produce less similar multimodal embeddings and when the subject in the image is sufficiently large.

Arguably, the more variance in classification score can be explained on the basis of such dataset variables, the less CLIP's benchmark score can be interpreted as an indicator of its linguistic understanding, thus calling into question the validity of the VALSE benchmark. 
However, none of the correlations found in the present study are particularly high and thus further analyses are needed to support this conclusion. 

\section{Limitations and future work}

The degree of localisation found in CLIP's text encoder is hard to interpret without reference to other results.
Future work could extend the present methodology to other tasks and potentially other models. 

Our study is also limited to simple effects of individual layer/position pairs. 
An analysis of the \emph{interaction} of certain layers or positions (e.g., by simultaneously patching activations in multiple places during causal tracing) might draw a more robust and conclusive picture of the inner processes that govern CLIP's understanding of negation. 

More generally, localisation methods may not be suited for analysing model behaviour that is shown with only moderate reliability. 
Note that the methods used in the present study had originally been proposed and applied to language model capabilities that are shown reliably across a large corpus of data, e.g., indirect object identification \citep{wang2022}, simple factual knowledge \citep{meng2023}, or docstring completion \citep{heimersheim2023}. 
By contrast, CLIP does not reliably handle negation in a multimodal context 
(CLIP's accuracy is only $66.9\%$) and these results are based on 
a relatively small dataset ($n=490$). 
In this case, methods like causal tracing do not intuitively lend themselves to \emph{comparing} situations evincing a particular model behaviour to those where the behaviour is absent.
That is because they focus on the degree to which an effect that represents a particular model behaviour can be restored or ablated, but 
the methodology breaks down when this effect isn't present in the first place. 

Thus, whilst illuminating the role of various components in CLIP's processing of negation, we cannot provide strong insights into why this processing yields correct classifications only in a fraction of cases. 
Furthermore, since correct classification only occurs in a subset of instances of VALSE, which 
is moderately sized 
to begin with, the results described here
require a larger and potentially more diverse dataset to obtain greater validity. 

With respect to the validity of the underlying VALSE benchmark,
it might be worthwhile to conduct a larger study on dataset features (e.g., image brightness, contrast, etc.) that correlate with benchmark performance. 
Comparisons with other VL benchmarks would further help putting these results into perspective. 
Such features that are predictive of benchmark performance limit the validity of linguistic benchmarks 
and highlight variables that should be controlled for in the creation of future benchmarks.


\bibliography{acl_latex}

\clearpage
\appendix

\section{Appendix} \label{sec:appendix}

\subsection{Preprocessing of VALSE instances}
\label{app:rephrasing}
Since caption and foil in the VALSE existence dataset differ only in the presence of the negator, they sometimes have a different number of tokens. 
Concretely, this is the case in ``bare plural'' sentences where there is no article or other qualifier in the non-negated sentence (e.g., ``There are tennis players.'' vs ``There are \emph{no} tennis players.'').
Identifying differences in how CLIP processes negated vs. non-negated labels is a core facet of the present study and such comparisons are greatly facilitated if caption and foil have the same number of tokens. 
Therefore, labels were rephrased to achieve equal sequence length by inserting the qualifier ``some'' into the non-negated plural sentences right before the subject.
For example, ``There are tennis players'' was rephrased to ``There are \emph{some} tennis players''. 
$15$ instances ($0.03\%$) from the original dataset have labels that do not follow the simple ``There is/are no [subject] ...'' structure and therefore aren't amenable to the rephrasing rule described above. 
For reasons of simplicity, these were omitted from the rephrased dataset. 

Importantly, rephrasing the dataset in this way only led to minor changes in CLIP's classification accuracy on this dataset ($0.691$ before, $0.686$ after rephrasing).
All analyses are based on the rephrased dataset, unless denoted otherwise. 

\subsection{CANNOT dataset}
\label{app:cannot}

We use the CANNOT dataset to indpendently validate our analysis of negator-selective attention in the CLIP text encoder. 

For the present purposes, the dataset is filtered to $554$ negated sentences that contain the word ``no'' as the determiner of the sentence subject (e.g., ``Medical organizations recommend \emph{no} alcohol during pregnancy for this reason''), using a tokeniser from the \texttt{spacy} Python library \citep{honnibal2020}.

For each of these sentences, a non-negated counterpart is then generated by replacing the word ``no'' with ``some''. 

This yields a set of sentence pairs, comparable to the caption-foil pairs from VALSE existence, which thus allows us to apply the same methodology for negator-selective attention. 

\subsection{Negator-selective attention on VALSE}
\label{app:appendix_attention}

\begin{figure*}
    \centering
    \includegraphics[width=\textwidth]{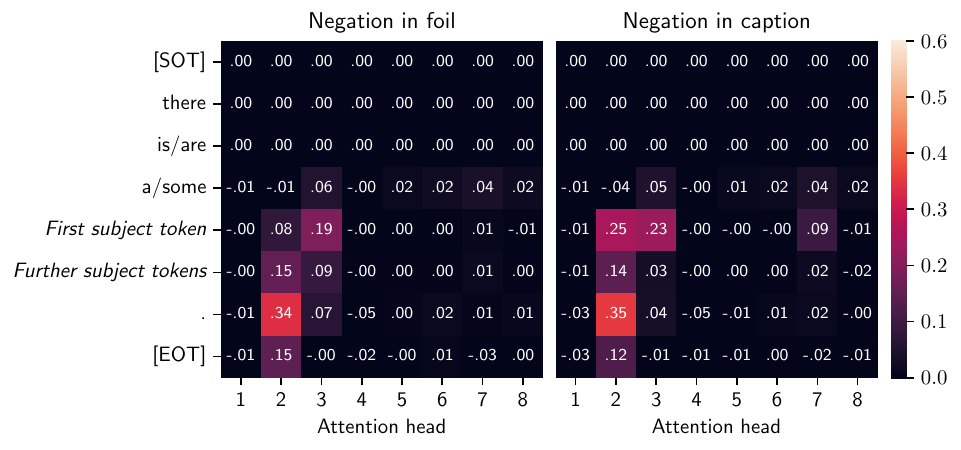}
    \caption[Source of negator-selective attention in layer 4]{
        Source of negator-selective attention in layer 4 across all dataset segments, split by whether negation is in foil or caption.
        The heatmaps show the degree of negator-selective attention from each sequence position (y-axis) in each attention head (x-axis).
        }
    \label{fig:attention_layer_4}
\end{figure*}

As discussed in Section \ref{sec:result_attention},
Figure \ref{fig:attention_layer_4} confirms that the source of negator-specific attention in Head 2 is the second-to-last position.

\subsection{Negator-selective attention results on CANNOT}
\label{app:cannot-results}

\begin{figure*}
    \centering
    \includegraphics[width=0.69\textwidth]{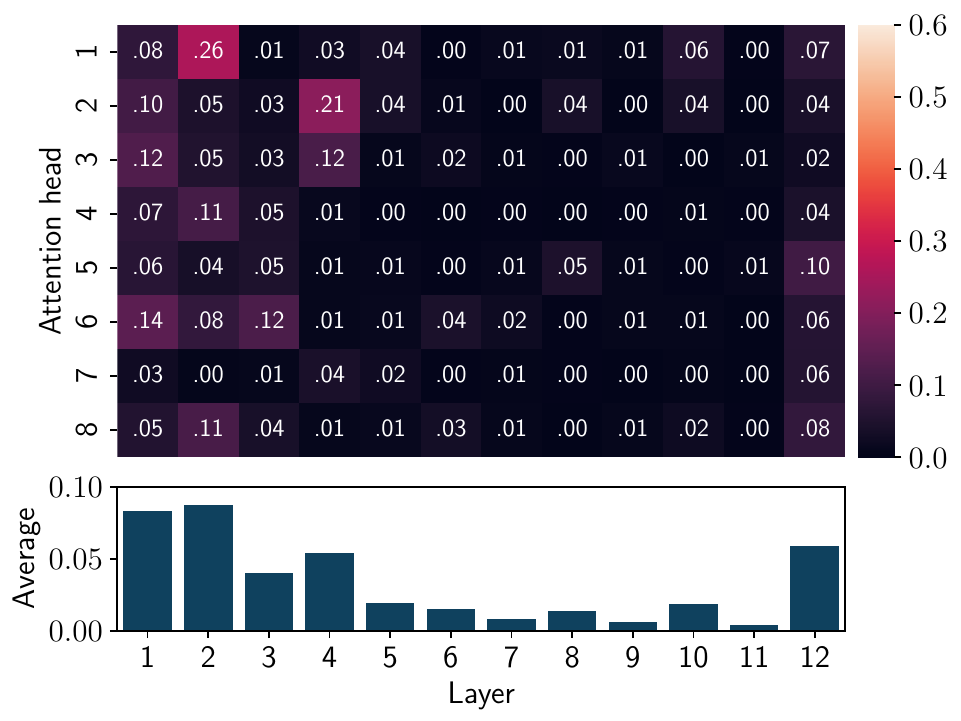}
    \caption[Validation of negator-selective attention on the CANNOT dataset]{
        Negator-selective attention on the CANNOT dataset, to validate the results from Figure~\ref{fig:attention_2in1}.
        The heatmaps indicate the degree of negator-selective attention for each attention head and layer.
        The bar charts show the average of each layer as an overall measure of negator-selective attention.
        }
    \label{fig:attention_cannot}
\end{figure*}

For validation purposes, Figure \ref{fig:attention_cannot} shows negator-selective attention on a subset of the CANNOT dataset.
Just like on the VALSE dataset, most negator-selective attention is found in the early layers~1-4. 
Head 2 in layer 4 once again shows particularly high negator-selective attention, albeit not the highest, which here is found in head 1 in layer 2. 
In summary, this provides converging evidence for the negator-selective attention results found in VALSE existence. 

\subsection{Dataset features}
\label{app:data_feat}

\begin{figure*}
    \centering
    \includegraphics[width=0.9\textwidth]{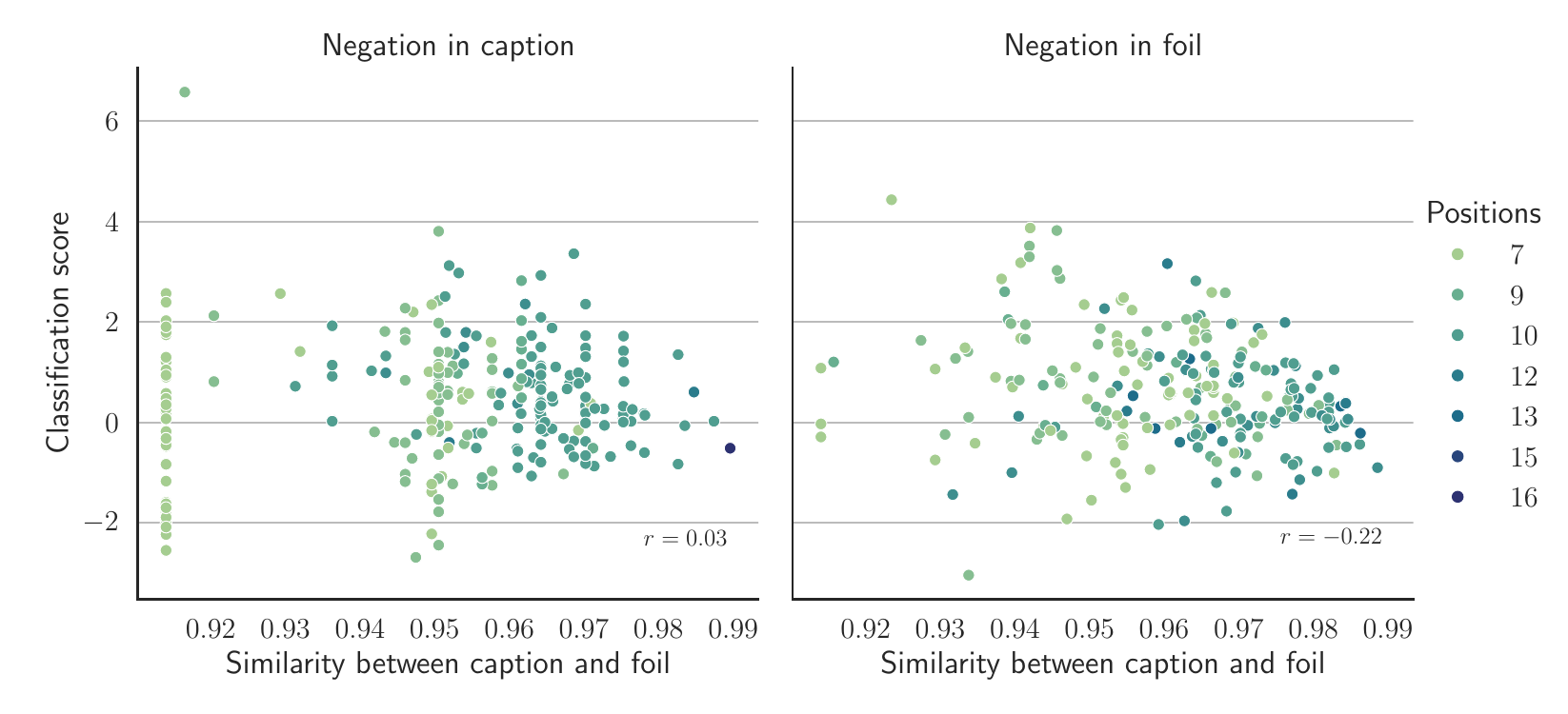}
    \caption[Caption-foil similarity vs. CLIP's classification score]{
        Cosine similarity of caption and foil in CLIP's multimodal embedding space vs. CLIP's classification score.
        Colour indicates dataset sequence length (i.e., number of tokens in sequence).
        }
    \label{fig:score_vs_similarity}
\end{figure*}

Figure \ref{fig:score_vs_similarity} shows the cosine similarity of each instance's caption and foil in CLIP's multimodal embedding space against that instance's classification score, split by whether the negation is in the caption or foil. 

When the negation is in the foil, similarity and score are weakly correlated ($r=-0.22$), whereas no correlation is found when the negation is in the caption ($r=0.03$). The latter is however influenced by the presence of a set of outliers, all with the same caption ``There are no people.''. Removing them from this analysis 
yields a correlation of $r=-0.20$, comparable to the one found when the negation is in the foil.

Figure \ref{fig:score_vs_similarity} also encodes sequence length, with longer sequences (darker colour) tending to exhibit greater caption-foil similarity. 
This is to be expected since caption and foil differ in exactly one position. 
If the total number of positions increases, then the relative size of this difference decreases, leading to greater similarity. 
These results suggest that CLIP's failure to correctly classify some VALSE Existence instances might be partly due to instances with longer captions and foils that are more similar in their representation and therefore more difficult to tell apart.
However, filtering the dataset to instances with shorter sequences does not meaningfully improve CLIP's accuracy, suggesting that sequence length plays a minor role at best. 

\end{document}